\documentclass[final]{cvpr}

\usepackage{times}
\usepackage{epsfig}
\usepackage{graphicx}
\usepackage{amsmath}
\usepackage{amssymb}

\usepackage{framed}
\usepackage{balance}
\usepackage{booktabs}
\usepackage{bm}
\usepackage{bbm}
\usepackage{xspace}
\usepackage{lipsum}
\usepackage{enumitem}
\usepackage{tablefootnote}
\usepackage[export]{adjustbox}
\usepackage{pifont}

\newcommand{\tinytit}[1]{\noindent\textbf{#1.}}

\usepackage[pagebackref=true,breaklinks=true,colorlinks,bookmarks=false]{hyperref}

\def\cvprPaperID{0010} % *** Enter the CVPR Paper ID here
\def\confYear{CVPRW 2021}
\begin{document}

%%%%%%%%% TITLE
\title{Revisiting The Evaluation of Class Activation Mapping for Explainability:\\A Novel Metric and Experimental Analysis}

\author{Samuele Poppi \quad Marcella Cornia \quad Lorenzo Baraldi \quad Rita Cucchiara \\
University of Modena and Reggio Emilia\\
{\tt\small 186923@studenti.unimore.it, \{marcella.cornia,lorenzo.baraldi,rita.cucchiara\}@unimore.it}
}

\maketitle

%%%%%%%%% ABSTRACT
\begin{abstract}
As the request for deep learning solutions increases, the need for explainability is even more fundamental. In this setting, particular attention has been given to visualization techniques, that try to attribute the right relevance to each input pixel with respect to the output of the network. In this paper, we focus on Class Activation Mapping (CAM) approaches, which provide an effective visualization by taking weighted averages of the activation maps. To enhance the evaluation and the reproducibility of such approaches, we propose a novel set of metrics to quantify explanation maps, which show better effectiveness and simplify comparisons between approaches. To evaluate the appropriateness of the proposal, we compare different CAM-based visualization methods on the entire ImageNet validation set, fostering proper comparisons and reproducibility.
\end{abstract}

\section{Introduction}
Explaining neural network predictions has been recently gaining a lot of attention in the research community, as it can increase the transparency of learned models and help to justify incorrect outputs in a human-friendly way. While there have been diverse attempts to provide explanations about the inference process in different forms~\cite{hendricks2016generating,hendricks2018grounding,goyal2019counterfactual}, the graphical visualization of a quantity of interest (\eg~regions of the input) remains the most straightforward and effective explanation approach.

\begin{figure}[t]
    \begin{center}
        \includegraphics[width=\linewidth]{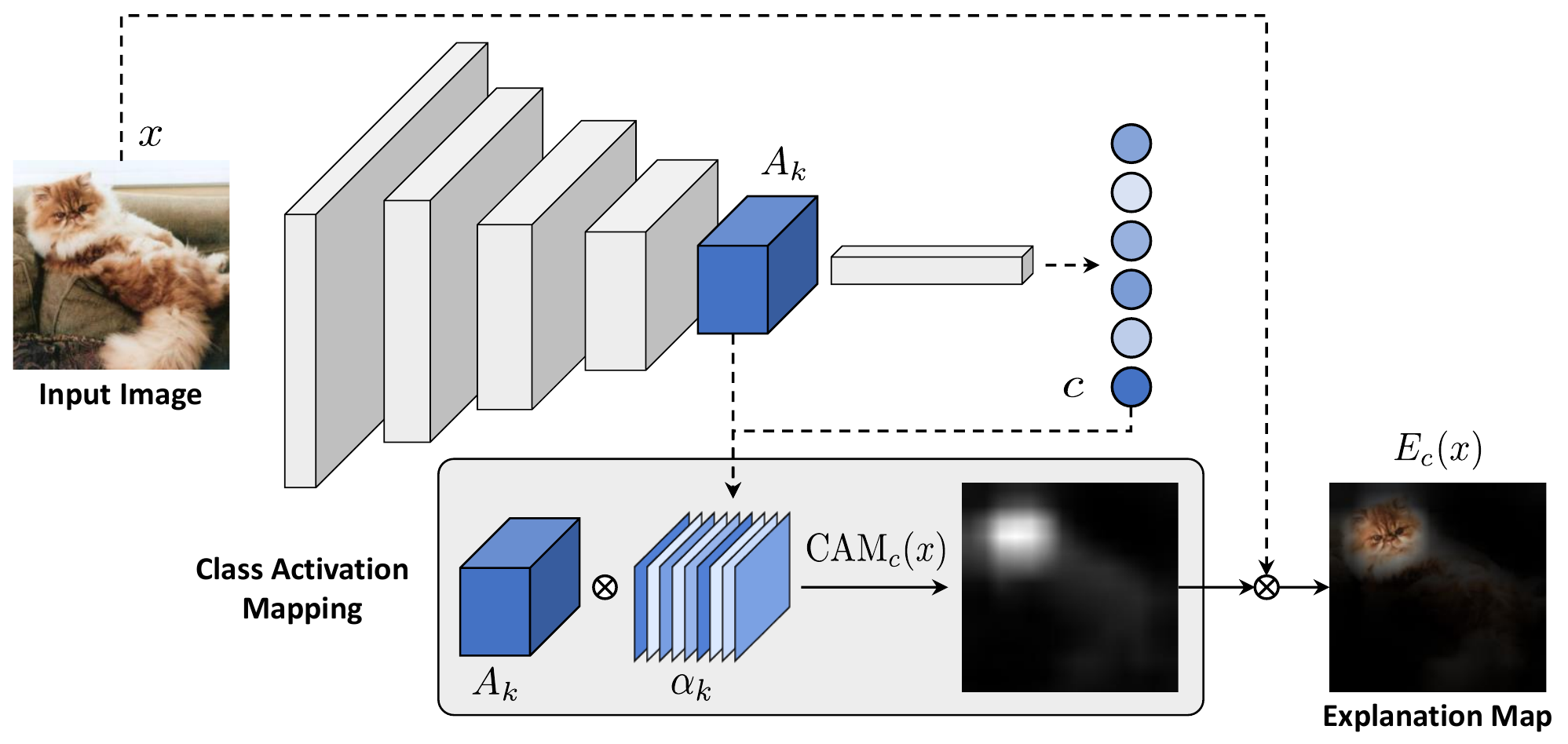}
    \end{center}
     \vspace{-.2cm}
\caption{Overview of CAM approaches for explaining predictions: explanation maps are produced via a linear combination of the activations of a convolutional layer.}
    \label{fig:first_page}
     \vspace{-.3cm}
\end{figure}

Because of the effectiveness of visualizations, there has been a surge of methods to solve the task, including gradient visualization tools~\cite{simonyan2013deep,zeiler2014visualizing}, gradient-based~\cite{springenberg2014striving,sundararajan2017axiomatic,lundberg2017unified,shrikumar2017learning,adebayo2018sanity}, and perturbation-based approaches~\cite{ribeiro2016should,dabkowski2017real,fong2017interpretable,petsiuk2018rise,chang2019explaining,wagner2019interpretable}. Among them, Class Activation Mapping (CAM)~\cite{zhou2016learning,selvaraju2017grad,chattopadhay2018grad,fu2020axiom,wang2020score,ramaswamy2020ablation,naidu2020ss,naidu2020cam} provides effective visual explanations by taking a weighted combination of activation maps from a convolutional layer. The motivation behind the approach is that each activation map contains different spatial information about the input, and when the selected convolutional layer is close to the classification stage of the network, its activations are sufficiently high-level to provide a visual localization that explains the final prediction. Identifying a proper way of calculating the importance (\ie,~the weight) of each channel is the main issue that has been tackled by all recent CAM approaches~\cite{selvaraju2017grad,fu2020axiom,wang2020score}.

As it often happens when a new field emerges, the comparison of different CAM approaches has been done mainly in a qualitative way, through the visual comparison of explanation maps, or via quantitative metrics which, however, are not completely effective and sometimes fail to numerically convey the quality of the explanation. At the same time, the evaluation has mostly been limited to few backbones and using protocols that imply a random selection of data and are not completely replicable. With the aim of improving the evaluation of CAM-based approaches, in this paper, we propose a novel set of metrics for CAM analysis, which provides a better ground for evaluation and simplifies comparisons. The effectiveness of the proposed metrics is assessed by comparing a variety of CAM-based approaches and by running experiments in a fully replicable setting.

\section{Preliminaries} \label{sec:method}
Let $f$ be a CNN-based classification model and $c$ a target class of interest. Given an input image $x$ and a convolutional layer of $f$, the \textit{Class Activation Mapping}~\cite{zhou2016learning} with respect to $c$ can be defined as a linear combination of the activation maps of the convolutional layer (Fig.~\ref{fig:first_page}), as follows:
\begin{equation}
    \text{CAM}_c(x) = \text{ReLU}\left( \sum_{k=1}^{N_l} \alpha_k A_k\right),
\end{equation}
where $N_l$ denotes the number of channels of the convolutional layer, $A_k$ is the $k$-th channel of the activation, and $\alpha_k$ are weight coefficients indicating the importance of the activation maps with respect to the target class. Depending on the specific CAM approach, these weights can be in scalar or matrix form, so that it is possible to apply a pixel-level weighting of the activation map. A $\text{ReLU}$ activation is employed to consider only the features that have a positive influence on the class of interest, \ie,~pixels whose intensity should be increased to increase the score for class $c$.

Regardless of the particular CAM approach at hand, $\text{CAM}_c(x)$ is usually upsampled to the size of the input image to obtain fine-grained pixel-scale representations. From this, an explanation map $\text{E}_c(x)$ can be generated by taking the element-wise multiplication between $\text{CAM}_c(x)$ and the input image itself (see Fig.~\ref{fig:first_page}).

The concept of CAM has been firstly defined in~\cite{zhou2016learning} for CNNs with a global average pooling layer after the last convolutional layer. In this case, weights $\alpha_k$ were defined as the weights of the final classification layer. Subsequently, several more sophisticated approaches for computing $\alpha_k$ have been proposed. Grad-CAM~\cite{selvaraju2017grad} generalizes~\cite{zhou2016learning} to be applied to any network architecture. It computes the gradient of the score for the target class with respect to the activation map and then applies a global average pooling. Recently, an axiom-based version~\cite{fu2020axiom} has been introduced to improve Grad-CAM's sensitivity~\cite{sundararajan2017axiomatic} and conservation~\cite{montavon2018methods}.
%Formally, in this case the weights $\alpha_k$ could be expressed as $\alpha_k = \texttt{wmean}(\partial y_c / \partial A_k, \mathbf{1})$, where $\texttt{wmean}$ indicates a weighted mean over the spatial axes, using a matrix filled with ones as weights.
 % , $y_c$,

Grad-CAM++~\cite{chattopadhay2018grad}, instead, takes a true weighted average of the gradients.
%In this case, the average weights are determined by the second-order derivative of the score of the target class with respect to activation maps. 
%Formally, $\alpha_k = \texttt{wmean}(\partial y_c / \partial A_k, w^c_k)$, where $w^c_k$ are scalar weights. 
Each weight of the average is in turn obtained as a weighted average of the partial derivatives along the spatial axes, so to capture the importance of each location of activation maps.
The approach has been further extended in~\cite{omeiza2019smooth} by adding a smoothening technique in the gradient computation.
Score-CAM~\cite{wang2020score}, finally, avoids the usage of gradients and instead computes the weights $\alpha_k$ using a channel-wise increase of confidence, computed as the difference in confidence when feeding the network with the input $x$ multiplied by $A_k$ and that of a baseline input.

\section{Evaluating CAMs}
Ideally, the explanation map produced by a CAM approach should contain the minimum set of pixels that are relevant to explain the network output. While this has been mainly qualitatively evaluated, a quantitative evaluation of explanation capabilities is still in an early stage, with the appearance of different evaluation metrics~\cite{chattopadhay2018grad,petsiuk2018rise,fu2020axiom}, which although has not been unified throughout the community. In particular, we focus on the following metrics which have been recently proposed, and which focus on the change of model confidence induced by the explanation map.

\tinytit{Average Drop}
It measures the average percentage drop in confidence for the target class $c$ when the model sees only the explanation map, instead of the full image. For one image, the metric is defined as $(\max(0, y_c - o_c) / y_c)\cdot 100$, where $y_c$ is the output score for class $c$ when using the full image, and $o_c$ the output score when using the explanation map. The value is then averaged over a set of images.

\tinytit{Average Increase}
It computes, instead, the number of times the confidence of the model is higher when using the explanation map compared to when using the entire image. Formally, for a single image it is defined as $\mathbbm{1}_{y_c < o_c} \cdot 100$, where $\mathbbm{1}$ is the indicator function. The value is then again averaged over different images.

\tinytit{Insertion and Deletion} Deletion measures the drop in the probability of the target class as important pixels (given by the CAM) are gradually removed from the image, while insertion computes the rise in the target class probability as pixels are added according to the CAM. Both metrics are expressed in terms of the total Area Under the Curve.

\begin{table}
\small
\setlength{\tabcolsep}{.3em}
\begin{center}
\resizebox{\linewidth}{!}{
\begin{tabular}{lc cc c cc}
\toprule
& & \multicolumn{2}{c}{\textbf{VGG-16}} & & \multicolumn{2}{c}{\textbf{ResNet-50}} \\
\cmidrule{3-4} \cmidrule{6-7}
\textbf{Method} & & Avg Drop $\downarrow$ & Avg Increase $\uparrow$ & & Avg Drop $\downarrow$ & Avg Increase $\uparrow$ \\
\midrule
Grad-CAM~\cite{selvaraju2017grad} & & 66.42 & 5.92 & & 32.99 & 24.27 \\
XGrad-CAM~\cite{fu2020axiom} & & 73.84 & 4.09 & & - & - \\
Grad-CAM++~\cite{chattopadhay2018grad} & & 32.88 & 20.10 & & 12.82 & 40.63 \\
Smooth Grad-CAM++~\cite{omeiza2019smooth} & & 36.72 & 16.11 & & 15.21 & 35.62 \\
Score-CAM~\cite{wang2020score} & & 26.13 & 24.75 & & 8.61 & 46.00 \\
\midrule
Fake-CAM & & \textbf{0.15} & \textbf{45.51} & & \textbf{0.38} & \textbf{47.54} \\
\bottomrule
\end{tabular}
}
\end{center}
\vspace{-.1cm}
\caption{Average Drop and Average Increase values of different CAM approaches, in comparison with Fake-CAM.}
\label{tab:limitations}
\vspace{-0.2cm}
\end{table}

\begin{figure*}[t]
\scriptsize
\begin{center}
\setlength{\tabcolsep}{.3em}
\resizebox{\linewidth}{!}{
\begin{tabular}{cccccc c ccccc}
\multicolumn{6}{c}{\footnotesize\textbf{VGG-16}} & & \multicolumn{5}{c}{\footnotesize\textbf{ResNet-50}} \\
\addlinespace[0.15cm]
\textbf{Image} & \cite{selvaraju2017grad} & \cite{fu2020axiom} & \cite{chattopadhay2018grad} & \cite{omeiza2019smooth} & \cite{wang2020score} & & \textbf{Image} & \cite{selvaraju2017grad} & \cite{chattopadhay2018grad} & \cite{omeiza2019smooth} & \cite{wang2020score} \\
\addlinespace[0.1cm]
\includegraphics[width=0.11\linewidth]{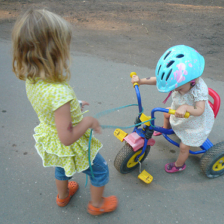} &
\includegraphics[width=0.11\linewidth]{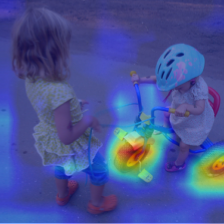} &
\includegraphics[width=0.11\linewidth]{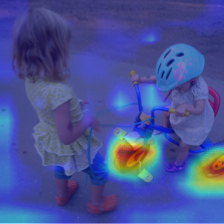} &
\includegraphics[width=0.11\linewidth]{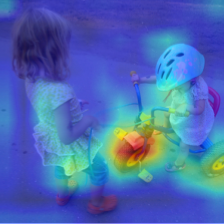} &
\includegraphics[width=0.11\linewidth]{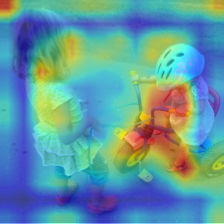} &
\includegraphics[width=0.11\linewidth]{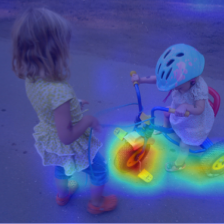} & & 
\includegraphics[width=0.11\linewidth]{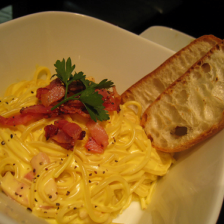} &
\includegraphics[width=0.11\linewidth]{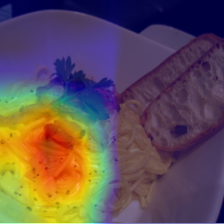} &
\includegraphics[width=0.11\linewidth]{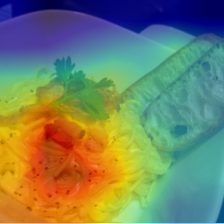} &
\includegraphics[width=0.11\linewidth]{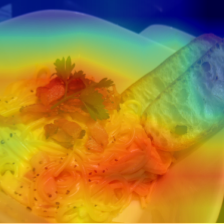} &
\includegraphics[width=0.11\linewidth]{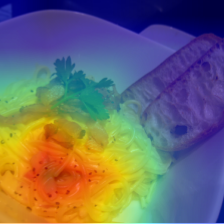} \\
\textbf{Class:} & 
\textbf{Avg Drop:}~93.26 & 
\textbf{Avg Drop:}~66.21 & 
\textbf{Avg Drop:}~59.92 & 
\textbf{Avg Drop:}~25.86 & 
\textbf{Avg Drop:}~38.51 & & 
\textbf{Class:} &
\textbf{Avg Drop:}~0.00 & 
\textbf{Avg Drop:}~0.00 & 
\textbf{Avg Drop:}~0.00 & 
\textbf{Avg Drop:}~0.00 \\
\textit{tricycle} & 
\textbf{Coherency:}~75.13 & 
\textbf{Coherency:}~74.65 & 
\textbf{Coherency:}~91.95 & 
\textbf{Coherency:}~80.50 & 
\textbf{Coherency:}~96.79 & &
\textit{carbonara} &
\textbf{Coherency:}~97.39 & 
\textbf{Coherency:}~98.55 & 
\textbf{Coherency:}~95.63 & 
\textbf{Coherency:}~98.41 \\
&
\textbf{Complexity:}~8.91 & 
\textbf{Complexity:}~8.98 & 
\textbf{Complexity:}~20.24 & 
\textbf{Complexity:}~37.66 & 
\textbf{Complexity:}~9.91 & 
& &
\textbf{Complexity:}~18.57 & 
\textbf{Complexity:}~43.43 & 
\textbf{Complexity:}~58.80 & 
\textbf{Complexity:}~34.30 \\
& 
\textbf{\textcolor{red}{ADCC:}}~17.36 & 
\textbf{\textcolor{red}{ADCC:}}~55.58 & 
\textbf{\textcolor{red}{ADCC:}}~62.03 & 
\textbf{\textcolor{red}{ADCC:}}~71.51 & 
\textbf{\textcolor{red}{ADCC:}}~79.59 & 
& &
\textbf{\textcolor{red}{ADCC:}}~92.17 & 
\textbf{\textcolor{red}{ADCC:}}~79.31 & 
\textbf{\textcolor{red}{ADCC:}}~67.07 & 
\textbf{\textcolor{red}{ADCC:}}~84.79 \\
\addlinespace[0.1cm]
\includegraphics[width=0.11\linewidth]{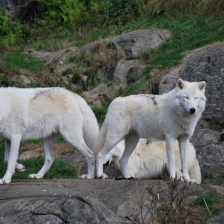} &
\includegraphics[width=0.11\linewidth]{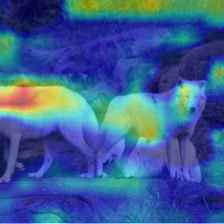} &
\includegraphics[width=0.11\linewidth]{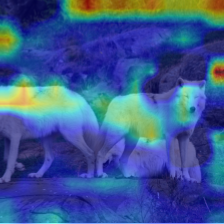} &
\includegraphics[width=0.11\linewidth]{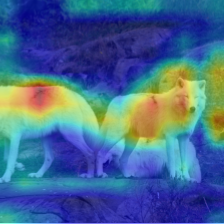} &
\includegraphics[width=0.11\linewidth]{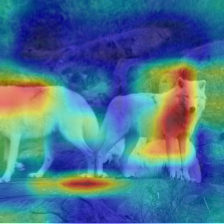} &
\includegraphics[width=0.11\linewidth]{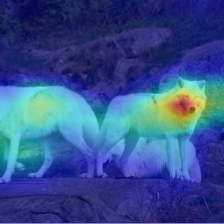} & & 
\includegraphics[width=0.11\linewidth]{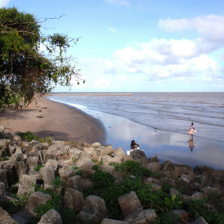} &
\includegraphics[width=0.11\linewidth]{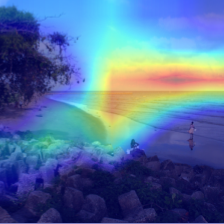} &
\includegraphics[width=0.11\linewidth]{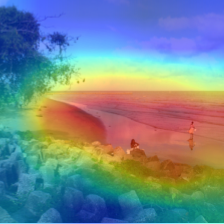} &
\includegraphics[width=0.11\linewidth]{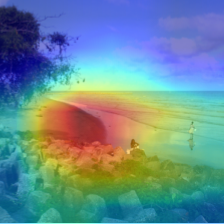} &
\includegraphics[width=0.11\linewidth]{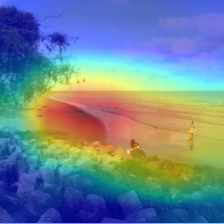} \\
\textbf{Class:} & 
\textbf{Avg Drop:}~53.52 & 
\textbf{Avg Drop:}~83.78 & 
\textbf{Avg Drop:}~0.00 & 
\textbf{Avg Drop:}~0.00 & 
\textbf{Avg Drop:}~0.05 & &
\textbf{Class:} &
\textbf{Avg Drop:}~0.00 & 
\textbf{Avg Drop:}~0.00 & 
\textbf{Avg Drop:}~0.00 & 
\textbf{Avg Drop:}~0.00 \\
\textit{white wolf} & 
\textbf{Coherency:}~73.41 & 
\textbf{Coherency:}~61.60 & 
\textbf{Coherency:}~95.24 & 
\textbf{Coherency:}~78.05 & 
\textbf{Coherency:}~95.14 & &
\textit{breakwater} &
\textbf{Coherency:}~65.69 & 
\textbf{Coherency:}~99.01 & 
\textbf{Coherency:}~96.43 & 
\textbf{Coherency:}~99.23 \\
&
\textbf{Complexity:}~21.79 & 
\textbf{Complexity:}~20.10 & 
\textbf{Complexity:}~28.19 & 
\textbf{Complexity:}~33.04 & 
\textbf{Complexity:}~14.92 & 
& &
\textbf{Complexity:}~21.51 & 
\textbf{Complexity:}~45.68 & 
\textbf{Complexity:}~37.59 & 
\textbf{Complexity:}~38.25 \\
& 
\textbf{\textcolor{red}{ADCC:}}~62.60 & 
\textbf{\textcolor{red}{ADCC:}}~33.18 & 
\textbf{\textcolor{red}{ADCC:}}~87.15 & 
\textbf{\textcolor{red}{ADCC:}}~79.48 & 
\textbf{\textcolor{red}{ADCC:}}~92.96 & 
& &
\textbf{\textcolor{red}{ADCC:}}~79.02 & 
\textbf{\textcolor{red}{ADCC:}}~77.91 & 
\textbf{\textcolor{red}{ADCC:}}~82.44 & 
\textbf{\textcolor{red}{ADCC:}}~82.71 \\
\end{tabular}
}
\end{center}
\vspace{-0.1cm}
\caption{Explanation maps and evaluation scores of different approaches on sample images from ImageNet validation set. We compare the results of Grad-CAM~\cite{selvaraju2017grad}, XGrad-CAM~\cite{fu2020axiom} (for VGG-16 only), Grad-CAM++~\cite{chattopadhay2018grad}, SmoothGrad-CAM++~\cite{omeiza2019smooth}, and ScoreCAM~\cite{wang2020score}.}
\label{fig:saliency}
\vspace{-0.25cm}
\end{figure*}

\subsection{Limitations}
While the rise of quantitative evaluation approaches is valuable for the field, many of the proposed metrics lack in providing a proper and affordable evaluation for explainability. From a numerical point of view, having a set of different metrics rather than a single-valued score makes comparison between different approaches cumbersome. Secondly, while average increase is too discrete for evaluating the rise of model confidence, average drop alone can easily bring to a misleading evaluation.

To further showcase the limitations of average drop and average increase, we build a ``fake'' CAM approach in which weights $\alpha_k$ do not depend on confidence scores. Specifically, for each activation map Fake-CAM produces a weight $\alpha_k$ in matrix form, in which all pixels are set to $1/N_l$, where $N_l$ is the number of activation maps, except for the top-left pixel, which is set to zero. The result is a class activation map which is 1 almost everywhere, except for the top-left pixel which is set to 0. 
Because the resulting explanation map is almost equivalent to the original image, except for one pixel in the corner which is unlikely to contain the target class, the average increase of Fake-CAM is usually very high. When computed on the entire ImageNet validation set~\cite{russakovsky2015imagenet} surpasses $45\%$ when employing most backbones -- a value that is superior to that of any true CAM approach (see Table~\ref{tab:limitations}). Similarly, the average drop of Fake-CAM is almost zero, because of the similarity between the input image and the explanation map. While Fake-CAM clearly does not help to explain the model predictions, it achieves almost ideal scores in terms of both Average Drop and Average Increase.

\subsection{Proposed Metrics}
In order to define a better evaluation protocol, we start by defining which properties an ideal attribution method should verify. The final proposed metric is a combination of three scores, each tackling one of the ideal properties.

\begin{table*}
\small
\setlength{\tabcolsep}{.25em}
\begin{center}
\resizebox{\linewidth}{!}{
\begin{tabular}{l ccccccc c ccccccc}
\toprule
& \multicolumn{7}{c}{\textbf{VGG-16}} & & \multicolumn{7}{c}{\textbf{ResNet-18}} \\
\cmidrule{2-8} \cmidrule{10-16}
\textbf{Method} & Avg Drop $\downarrow$ & Avg Inc $\uparrow$ & Deletion $\downarrow$ & Insertion $\uparrow$ & \textbf{Coherency} $\uparrow$ & \textbf{Complexity} $\downarrow$ & \textbf{ADCC} $\uparrow$ & & Avg Drop $\downarrow$ & Avg Inc $\uparrow$ & Deletion $\downarrow$ & Insertion $\uparrow$ & \textbf{Coherency} $\uparrow$ & \textbf{Complexity} $\downarrow$ & \textbf{ADCC} $\uparrow$ \\
\midrule
Fake-CAM & \textit{0.15} & \textit{45.51} & \textit{32.87} & \textit{35.70} & \textit{100.00} & \textit{100.00} & \textit{0.01} & & \textit{0.24} & \textit{45.37} & \textit{31.12} & \textit{33.44} & \textit{100.00} & \textit{100.00} & \textit{0.01} \\
\midrule
Grad-CAM~\cite{selvaraju2017grad} & 66.42 & 5.92 & 11.12 & 19.56 & 69.20 & 15.65 & 53.52 & & 42.90 & 16.63 & 13.43 & 41.47 & 81.03 & \textbf{23.04} & 69.98 \\
XGrad-CAM~\cite{fu2020axiom} & 73.84 & 4.09 & 11.59 & 14.95 & 66.69 & \textbf{13.68} & 46.29 & & - & - & - & - & - & - & - \\
Grad-CAM++~\cite{chattopadhay2018grad} & 32.88 & 20.10 & \textbf{8.82} & 36.60 & 89.34 & 26.33 & 75.65 & & 17.85 & 34.46 & 12.30 & 44.80 & 98.18 & 44.63 & 74.24 \\
Smooth Grad-CAM++~\cite{omeiza2019smooth} & 36.72 & 16.11 & 10.57 & 31.36 & 82.68 & 28.09 & 71.72 & & 20.67 & 29.99 & 12.83 & 43.13 & 97.53 & 43.11 & 74.20 \\
Score-CAM~\cite{wang2020score} & \textbf{26.13} & \textbf{24.75} & 9.52 & \textbf{47.00} & \textbf{93.83} & 20.27 & \textbf{81.66} & & \textbf{12.81} & \textbf{40.41} & \textbf{10.76} & \textbf{46.01} & \textbf{98.35} & 41.78 & \textbf{77.30} \\
% Smoothed Score-CAM~\cite{} & . & . & . & . & . & . & . & & . & . & . & . & . & . & . \\
% Integrated Score-CAM~\cite{} & . & . & . & . & . & . & . & & . & . & . & . & . & . & . \\
\midrule
& \multicolumn{7}{c}{\textbf{ResNet-50}} & & \multicolumn{7}{c}{\textbf{ResNet-101}} \\
\cmidrule{2-8} \cmidrule{10-16}
\textbf{Method} & Avg Drop $\downarrow$ & Avg Inc $\uparrow$ & Deletion $\downarrow$ & Insertion $\uparrow$ & \textbf{Coherency} $\uparrow$ & \textbf{Complexity} $\downarrow$ & \textbf{ADCC} $\uparrow$ & & Avg Drop $\downarrow$ & Avg Inc $\uparrow$ & Deletion $\downarrow$ & Insertion $\uparrow$ & \textbf{Coherency} $\uparrow$ & \textbf{Complexity} $\downarrow$ & \textbf{ADCC} $\uparrow$ \\
\midrule
Fake-CAM & \textit{0.38} & \textit{47.54} & \textit{38.06} & \textit{38.72} & \textit{100.00} & \textit{100.00} & \textit{0.01} & & \textit{0.36} & \textit{43.98} & \textit{43.66} & \textit{41.64} & \textit{100.00} & \textit{100.00} & \textit{0.01} \\
\midrule
Grad-CAM~\cite{selvaraju2017grad} & 32.99 & 24.27 & 17.49 & 48.48 & 82.80 & \textbf{22.24} & 75.27 & & 29.38 & 29.35 & 18.66 & 47.47 & 81.97 & \textbf{22.51} & 76.40 \\
% XGrad-CAM~\cite{fu2020axiom} & . & . & . & . & . & . & . & & . & . & . & . & . & . & . \\
Grad-CAM++~\cite{chattopadhay2018grad} & 12.82 & 40.63 & 14.10 & 53.51 & 97.84 & 43.99 & 75.86 & & 11.38 &  42.07 & 14.99 & 56.65 & 98.28 & 43.94 & 76.34 \\
Smooth Grad-CAM++~\cite{omeiza2019smooth} & 15.21 & 35.62 & 15.21 & 52.43 & 97.47 & 42.25 & 76.19 & & 13.37 & 37.76 & \textbf{14.32} & 58.23 & 97.76 & 42.61 & 76.54  \\
Score-CAM~\cite{wang2020score} & \textbf{8.61} & \textbf{46.00} & \textbf{13.33} & \textbf{54.16} & \textbf{98.12} & 42.05 & \textbf{78.14} & & \textbf{7.20} & \textbf{47.93} & 14.63 & \textbf{59.57} & \textbf{98.37} & 42.04 & \textbf{78.55} \\
% Smoothed Score-CAM~\cite{} & . & . & . & . & . & . & . & & . & . & . & . & . & . & . \\
% Integrated Score-CAM~\cite{} & . & . & . & . & . & . & . & & . & . & . & . & . & . & . \\
\midrule
& \multicolumn{7}{c}{\textbf{ResNeXt-50}} & & \multicolumn{7}{c}{\textbf{ResNeXt-101}} \\
\cmidrule{2-8} \cmidrule{10-16}
\textbf{Method} & Avg Drop $\downarrow$ & Avg Inc $\uparrow$ & Deletion $\downarrow$ & Insertion $\uparrow$ & \textbf{Coherency} $\uparrow$ & \textbf{Complexity} $\downarrow$ & \textbf{ADCC} $\uparrow$ & & Avg Drop $\downarrow$ & Avg Inc $\uparrow$ & Deletion $\downarrow$ & Insertion $\uparrow$ & \textbf{Coherency} $\uparrow$ & \textbf{Complexity} $\downarrow$ & \textbf{ADCC} $\uparrow$ \\
\midrule
Fake-CAM & \textit{0.34} & \textit{46.70} & \textit{41.67} & \textit{43.31} & \textit{100.00} & \textit{100.00} & \textit{0.01} & & \textit{0.26} & \textit{42.43} & \textit{48.90} & \textit{46.79} & \textit{100.00} & \textit{100.00} & \textit{0.01} \\
\midrule
Grad-CAM~\cite{selvaraju2017grad} & 28.06 & 29.42 & 20.73 & 50.30 & 82.72 & \textbf{25.57} & \textbf{76.09} & & 24.12 & 36.37 & 20.47 & 61.04 & 82.94 & \textbf{25.45} & \textbf{77.62} \\
% XGrad-CAM~\cite{fu2020axiom} & . & . & . & . & . & . & . & & . & . & . & . & . & . & . \\
Grad-CAM++~\cite{chattopadhay2018grad} & 11.12 & 41.38 & 17.07 & 56.05 & 97.30 & 48.66 & 73.16 & & 9.74 & 42.63 & 17.63 & 62.90 & 95.05 & 46.27 & 74.61 \\
Smooth Grad-CAM++~\cite{omeiza2019smooth} & 12.70 & 36.58 & 16.90 & 56.76 & 97.32 & 47.48 & 73.58 & & 9.49 & 40.43 & 17.67 & \textbf{64.16} & 96.81 & 49.24 & 73.03 \\
Score-CAM~\cite{wang2020score} & \textbf{7.20} & \textbf{45.70} & \textbf{15.59} & \textbf{57.92} & \textbf{98.00} & 46.86 & 75.38 & & \textbf{5.37} & \textbf{47.70} & \textbf{17.30} & 63.61 & \textbf{97.03} & 46.83 & 75.60 \\
% Smoothed Score-CAM~\cite{} & . & . & . & . & . & . & . & & . & . & . & . & . & . & . \\
% Integrated Score-CAM~\cite{} & . & . & . & . & . & . & . & & . & . & . & . & . & . & . \\
\bottomrule
\end{tabular}
}
\end{center}
\vspace{-.1cm}
\caption{Evaluation of different CAM-based approaches with existing and proposed metrics, on six different backbones.}
\label{tab:results}
\vspace{-0.3cm}
\end{table*}

\tinytit{Maximum Coherency}
The CAM should contain all the relevant features that explain a prediction and should remove useless features in a coherent way. As a consequence, given an input image $x$ and a class of interest $c$, the CAM of $x$ should not change when conditioning $x$ on the CAM itself. Formally,
\begin{equation}
    \text{CAM}_c(x \odot \text{CAM}_c(x)) = \text{CAM}_c(x).
    \label{eq:coherency_crit}
\end{equation}
Notice that this is equivalent to requiring that the CAM of one image should be equal to that of the explanation map obtained with the same CAM approach. To measure the extent to which an approach satisfies the coherency property, we define a metric that measures how much the CAM changes when smoothing pixels with a low attribution score. Following previous works in the comparison of saliency maps~\cite{riche2013saliency,cornia2016multi,cornia2017visual,bylinskii2018different,cornia2018predicting,cornia2018sam}, we use the Pearson Correlation Coefficient between the two CAMs considered in Eq.~\ref{eq:coherency_crit}:
\begin{equation}
    \text{Coherency}(x) = \frac{\mathrm{Cov}(\text{CAM}_c(x \odot \text{CAM}_c(x)), \text{CAM}_c(x))}{\sigma_{\text{CAM}_c(x \odot \text{CAM}_c(x))}\sigma_{\text{CAM}_c(x)}},
\end{equation}
where $\mathrm{Cov}$ indicates the covariance between two maps, and $\sigma$ the standard deviation. Since the Pearson Correlation Coefficient ranges between $-1$ and $1$, we normalize the Coherency score between $0$ and $1$ and, following existing metrics, we also define it as a percentage. Clearly, Coherency is maximized when the attribution method is invariant to change in the input image. 

\tinytit{Minimum Complexity}
Beyond requiring that the CAM should be coherent in removing features from the input image, we must also require it to be as less complex as possible, \ie,~it must contain the minimum set of pixels that explains the prediction. Employing the $L_1$ norm as a proxy of the complexity of a CAM, we define the Complexity measure as:
\begin{equation}
    \text{Complexity}(x) = \| \text{CAM}_c(x) \|_1.
\end{equation}
Complexity is minimized when the number of pixels highlighted by the attribution method is low.

\tinytit{Minimum Confidence Drop} An ideal explanation map should produce the smallest drop in confidence with respect to using the original input image. To express this third property, we directly employ the Average Drop metric, which linearly computes the drop in confidence.

\tinytit{Average DCC}
Finally, we combine the three scores in a single metric, which we name Average DCC, by taking their harmonic mean, as follows:
\begin{multline}
    \text{ADCC}(x) = 3  \left(
    \frac{1}{\text{Coherency}(x)} + \right. \\ 
    \left. + \frac{1}{1-\text{Complexity}(x)} +
    \frac{1}{1-\text{AverageDrop}(x)}
    \right) ^{-1}
\end{multline}

Compared to the usage of separate metrics as done in the past, Average DCC has the additional merit of being a single-valued metric with which direct comparisons between approaches are feasible. From a methodological point of view, instead, it takes into account the complexity of the explanation map as well as the coherency of the CAM approach to be evaluated.

\section{Experiments} \label{sec:experiments}

\tinytit{Experimental Setup}
Differently from previous works which conducted the evaluation on randomly selected images, we conduct experiments on the entire ImageNet validation set (ILSVRC2012)~\cite{russakovsky2015imagenet} consisting of $50\,000$ images, each representing one of the $1\,000$ possible object classes. To increase the generality of the evaluation, we use six different CNNs for object classification -- \ie,~VGG-16~\cite{simonyan2015very}, ResNet-18, ResNet-50, ResNet-101~\cite{he2016deep}, ResNeXt-50, and ResNeXt-101~\cite{xie2017aggregated}, applying each CAM approach on the last convolutional layer. According to the original paper~\cite{fu2020axiom}, XGrad-CAM is equivalent to Grad-CAM when applied to ResNet models and for this reason, we report the results of this method only on VGG-16. All images are resized and center cropped to $224 \times 224$, using mean and standard deviation values computed over the ImageNet training set to normalize the results.

\tinytit{Experimental Results}
Fig.~\ref{fig:saliency} shows the scores obtained by the proposed metrics on some sample images, using both VGG-16 and ResNet-50. As it can be seen, the three metrics are complementary in evaluating explanation maps and are equally weighted in the final ADCC score. For instance, in the top-left example, the map produced by Score-CAM~\cite{wang2020score} achieves the best ADCC score, as it performs favorably in terms of reduced confidence drop while maintaining high levels of coherency and being less complex than other maps. On the contrary, the map produced by SmoothGrad-CAM++~\cite{omeiza2019smooth} has a lower drop in confidence, but it is less coherent and more complex. 
Turning to the \textit{carbonara} example, the maps reported by all approaches have the same drop in confidence (\ie,~0), and similar coherency values. The map produced by Grad-CAM~\cite{selvaraju2017grad} has the lowest complexity, thus obtaining the best final score.

In Table~\ref{tab:results} we report the values obtained with existing and proposed metrics on all six backbones, on the entire ImageNet validation set. As it can be seen, the results of Fake-CAM are better than true CAM approaches on average drop and average increase for all considered backbones. On the contrary, the proposed ADCC score correctly penalizes the complexity of explanation maps generated by Fake-CAM, thus confirming the appropriateness of the proposed evaluation metric. Comparing true CAM methods, generally Score-CAM~\cite{wang2020score} achieves the best results on almost all metrics, except for complexity. It shall be noted, also, that Score-CAM~\cite{wang2020score} performs favorably on VGG and ResNet backbones, while Grad-CAM~\cite{selvaraju2017grad} achieves the best ADCC score when employing ResNeXt models -- something which was never tested in literature before. In terms of complexity, Grad-CAM~\cite{selvaraju2017grad} produces less complex but less coherent maps, and with higher confidence drops. The ADCC score jointly accounts for all three properties, providing a single-valued metric from which different approaches can be easily compared.

\section{Conclusion}
We presented a novel evaluation protocol for CAM-based explanation approaches. The proposed ADCC score takes into account the variation of model confidence, the coherency, and the complexity of explanation maps in a single score, providing an effective mean of comparison. Experiments have been conducted on the entire ImageNet validation set, with six different CNN backbones, testifying the appropriateness of the proposed score and its generality across different settings. 

% \balance
{\small
\bibliographystyle{ieee_fullname}
\bibliography{egbib}
}

\end{document}